\documentclass[11pt,a4paper]{article}
\usepackage[hyperref]{eacl2021}
\usepackage{times}
\usepackage{latexsym}
\usepackage{graphicx}

\usepackage{microtype}

\aclfinalcopy % Uncomment this line for the final submission
\title{Methods for the Design and Evaluation of HCI+NLP Systems}

\author{Hendrik Heuer \\
  Institute for Information Management \\
  University of Bremen \\
  Bremen, Germany \\
  \texttt{hheuer@uni-bremen.de} \\\And
  Daniel Buschek \\
  Department of Computer Science \\
  University of Bayreuth \\
  Bayreuth, Germany \\
  \texttt{daniel.buschek@uni-bayreuth.de} \\}

\date{}

\begin{document}
\maketitle
\begin{abstract}
HCI and NLP traditionally focus on different evaluation methods. While HCI involves a small number of people directly and deeply, NLP traditionally relies on standardized benchmark evaluations that involve a larger number of people indirectly. We present five methodological proposals at the intersection of HCI and NLP and situate them in the context of ML-based NLP models. Our goal is to foster interdisciplinary collaboration and progress in both fields by emphasizing what the fields can learn from each other.
\end{abstract}

\section{Introduction}

NLP is the subset of AI that is focused on the scientific study of linguistic phenomena \cite{acl_what_2021}. Human-computer interaction (HCI) is ``the study and practice of the design, implementation, use, and evaluation of interactive computing systems'' \cite{10.5555/3019322}. Grudin described HCI and AI as two fields divided by a common focus \cite{grudin2009ai}: While both are concerned with intelligent behavior, the two fields have different priorities, methods, and assessment approaches. In 2009, Grudin argued that while AI research traditionally focused on long-term projects running on expensive systems, HCI is focused on short-term projects running on commodity hardware. For successful HCI+NLP applications, a synthesis of both approaches is necessary. As a first step towards this goal, this article, informed by our sensibility as HCI researchers, provides five concrete methods from HCI to study the design, implementation, use, and evaluation of HCI+NLP systems.

%HCI and NLP traditionally rely on different evaluation methods.
%
One promising pathway for fostering interdisciplinary collaboration and progress in both fields is to ask what each field can learn from the methods of the other. On the one hand, while HCI directly and deeply involves the end-users of a system, NLP involves people as providers of training data or as judges of the output of the system. On the other hand, NLP has a rich history of standardised evaluation metrics with freely available datasets and comparable benchmarks. HCI methods that enable deep involvement are needed to better understand the perspective of people using NLP, or being affected by it, their experiences, as well as related challenges and benefits. 

As a synthesis of this user focus and the standardized benchmarks, HCI+NLP systems could combine more standardized evaluation procedures and material (data, tasks, metrics) with user involvement. This could lead to better comparability and clearer measures of progress. This may also spur systematic work towards ``grand challenges'', that is, uniting HCI researchers under a common goal~\cite{Kostakos2015}. 

To facilitate a productive collaboration between HCI+NLP, clearly defined tasks that attract a large number of researchers would be helpful. These tasks could be accompanied with data to train models, as a methodological approach from NLP, and methodological recommendations on how to evaluate these systems, as a methodological approach from HCI. One task could e.g. define which questions should be posed to experiment participants. If the questions regarding the evaluation of an experiment are fixed, the results of different experiments could be more comparable. This would not only unite a variety of research results, but it could also increase the visibility of the researchers who participate. Complementary, NLP could benefit from asking further questions about use cases and usage contexts, and from subsequently evaluating contributions in situ, including use by the intended target group (or indirectly affected groups) of NLP.

\begin{table}[]
\small
\begin{tabular}{l p{4cm} }
\hline
\textbf{Method} & \textbf{Description} \\
\hline
1. User-Centered NLP & user studies ensure that users understand the output and the explanations of the NLP system \\
2. Co-Creating NLP & deep involvement from the start enables users to actively shape a system and the problem that the system is solving \\
3. Experience Sampling & richer data collected by (active) users enables a deeper understanding of the context and the process in which certain data was created \\
4. Crowdsourcing & an evaluation at scale with humans-in-the-loop ensures high system performance and could prevent biased results or discrimination \\
5. User Models & simulating real users computationally can automate routine evaluation tasks to speed up the development \\
\hline
\end{tabular}
\caption{The five methodological proposals for HCI+ML that we present in this paper.}\label{fig:table}
\end{table}

In conclusion, both fields stand to gain an enriched set of methodological procedures, practices, and tools. In the following, we propose five HCI+NLP methods that we consider useful in advancing research in both fields. Table~1 provides a short description of each of the five HCI+NLP methods that this paper highlights. With our non-exhaustive overview, we hope to inspire interdisciplinary discussions and collaborations, ultimately leading to better interactive NLP systems -- both ``better'' in terms of NLP capabilities and regarding usability, user experience, and relevance for people.

\section{Methods For HCI+NLP}

This section presents and discusses a set of concrete ideas and directions for developing evaluation methods at the intersection of HCI and NLP.

\subsection{User-Centred NLP}

Our experience as researchers at the intersection of HCI+AI taught us that systems that may work from an AI perspective, may not be helpful to users. One example of this is an unpublished machine learning-based fake news detection based on text style. Even though this worked in principle with F1-scores of 80 and higher, pilot studies showed that the style-based explanations are not meaningful to users. Even for educated participants, it may be an overextension to comprehend such explanations about an ML-based system. This relates to previous work that showed an explanatory gap between what is available to explain ML-based systems and what users need to understand such systems \cite{elib_4444}. Far too frequently, NLP systems are built on assumptions about users, not based on insights about users. We argue that all ML systems aimed at users need to be evaluated with users. Following ISO 9241-210, user-centered design is an iterative process that involves repeatedly 1. specifying the context of use, 2. specifying requirements, 3. developing solutions, and 4. evaluating solutions, all in close collaboration with users \cite{polska2011ergonomics}.

Our review of prior work indicates that HCI and NLP follow different approaches regarding the requirements analysis and the evaluation of complex information systems. To the best of our knowledge, we did not find good examples for true interdisciplinary collaborations that contribute to both fields. While there are HCI contributions that leverage NLP technology, they rarely make a fundamental contribution towards computational linguistics, merely applying existing approaches. On the other hand, where NLP aims to make a contribution to an HCI-related field, this contribution is commonly presented without empirical evidence in the form of user studies. Our most fundamental and important contribution in this position paper is a call to recenter efforts in natural language processing around users. We argue that empirical studies with and of users are central to successful HCI+AL applications. A contribution on a system for recognizing fake news, for example, has to empirically show that the way the system predicts its results is helpful to users. Training an ML-based system with good intentions is not enough for real progress.

\subsection{Co-Creating NLP Systems}

While user-centered design is already a great improvement from developing systems based on assumptions, HCI has moved beyond it, involving users much deeper. With so-called Co-Creation, users are not just objects that are studied to build better systems, but subjects that actively shape the system. We, therefore, argue that HCI+NLP researchers should (co)-create services with users. \citet{jarke2021co}, among others, describes co-creation as a joint problem-making and problem-solving of researcher and user. This deep involvement of users enables novel ways of sharing expertise and control over design decisions.

Prior research showed how challenging it can be for users to understand complex, machine-learning-based systems like the recommendation system on YouTube \cite{10.1145/3415192}. The field of HCI, therefore, recognized the importance of involving users in the design, implementation, and evaluation of interactive computing systems. While users are frequently the subject of investigation, recent trends in interaction design aim to involve users much earlier and deeper.

If users are deeply involved in the design and development of NLP systems, they can share their expertise on the task at hand. On the one hand, this can yield insights into UI and interaction design for the NLP system~\cite{Yang2019}. On the other hand, it is relevant regarding the output. Sharing control is also crucial considering the potential biases enacted by such systems. Deep involvement of a diverse set of users could help prevent problematic applications of machine learning and prevent discrimination based on gender \cite{10.5555/3157382.3157584} or ethnicity \cite{pmlr-v81-buolamwini18a}.

\subsection{Collecting Context-Rich Text Data with the Experience Sampling Method (ESM)}
The need for very large text datasets in NLP has motivated and favored certain methods for data collection, such as scraping text from the web.
These methods assume that text is ``already there'', i.e. they do not consider or facilitate its creation: For example, scraping Wikipedia neither supports Wikipedia authors, nor does it care if authors would want to have their texts included in such models, or not. 

To advance future HCI+NLP applications, it could be helpful to create and deploy tools for more interactive data collection. One important method here is the \textit{experience sampling method (ESM)}~\cite{csikszentmihalyi2014,Berkel2017ExperienceSamplingReview}, which is used widely in HCI and could be deployed for NLP as well. This method of data collection repeatedly asks short questions throughout participants' daily lives, and thus captures data in context: For instance, an ESM smartphone app could prompt users to describe their current environment, an experience they had today, or to ``donate'' input and language data (e.g. from messaging) in an anonymous way~\cite{Bemmann2020, Buschek2018}. This could be enriched with further context (e.g. location, date, time, weather, phone sensors) to answer novel research questions, such as how a language model for a chatbot can improve its text generation and understanding by making use of the location or other context data. One important example for such experience sampling is work on citizen sociolinguistics, which explores how citizens can participate (often through mobile technologies) in sociolinguistic inquiry \cite{rymes2014citizen}.

% or a word processor plugin to ``donate'' the text the user is currently working on
%
Although it would be challenging to collect massive amounts of text using this method, the ESM-based data collection could be used to complement data collected via scarping (e.g. via finetuning with ESM data). ESM also supports more personalized and context-rich language data and models, from specific communities or contexts. This might cater to novel research questions, e.g. on context-based and personalized language modeling. More generally, methods like ESM furthermore give the people that act as data sources more of a ``say'' in the data collection for NLP, for instance, via explicitly sharing data via an interactive ESM application, or via their rich daily contexts being better represented in metadata.

\subsection{Involving the Crowd for Interactive Benchmark Evaluations}
As described, NLP has a strong tradition in using and reusing benchmark datasets, which are beneficial for comparable and standardized evaluations. However, some aspects cannot be evaluated in this way. First, comparisons with human language understanding or generation are limited to the (few) humans that originally provided data for the limited set of examples that these people had been given. Yet language understanding and use change over time, and vary between people and their backgrounds and contexts.
Second, ``offline'' evaluations without people cannot assess \textit{interactive use} of NLP systems by people (e.g. chatting with a bot, writing with AI text suggestions). Therefore, at the intersection of HCI and NLP, one may ask: Is it possible to keep the benefits of (large) standardized benchmark evaluations while involving humans?

\textit{Crowd-sourcing} may provide one approach to address this: HCI and NLP researchers should create evaluation tools that streamline large-scale evaluations with remote participants. Practically speaking, one would then still set a benchmark task running ``with one click'', yet this would trigger the creation, distribution, and collection of crowd-tasks. One example of this is ``GENIE'', a system and leaderboard for human-in-the-loop evaluation of text generation~\cite{khashabi2021genie}.

\subsection{Employing User Models as Proxies for Interactive Evaluations}

In addition to involving users deeply and collecting context-rich data, relevant aspects of people's interaction behavior with interactive NLP systems may also be modeled explicitly. HCI, psychology, and related fields offer a variety of models, for example, relating to pointing at user interface targets or selecting elements from a list. Extending and improving those modeled aspects is particularly pursued in the emerging area of \textit{Computational HCI}~\cite{oulasvirta2018computational}. Even though such models cannot replace humans, they may help evaluate certain aspects and parameter choices of an interactive NLP system in a standardized and rapid manner.

For instance, \citet{todi2021adapting} showed that approaches based on reinforcement learning can be used to automatically adapt related user interfaces. For interactive NLP, \citet{Buschek2021chi} investigated how different numbers of phrase suggestions from a neural language model impact user behavior while writing, collecting a dataset of 156 people's interactions. In the future, data such as this might be used, for example, to train a model that replicates users' selection strategies for text suggestions from an NLP system. Such a model might then be used in lieu of actual users to gauge general usage patterns for HCI+NLP systems, e.g. for interactive text generation.

%Overall, this seems to be the technically most challenging direction out of the three presented here. Potential technical approaches here might involve reinforcement learning (of an agent representing a user). Beyond user modeling, we see a strong overlap with work on intelligent and adaptive user interfaces, which are also interested in actionable expectations of users' interaction behavior (e.g. see~\cite{todi2021adapting}).

\section{Discussion}

\begin{figure}
\centering
\includegraphics[width=\columnwidth]{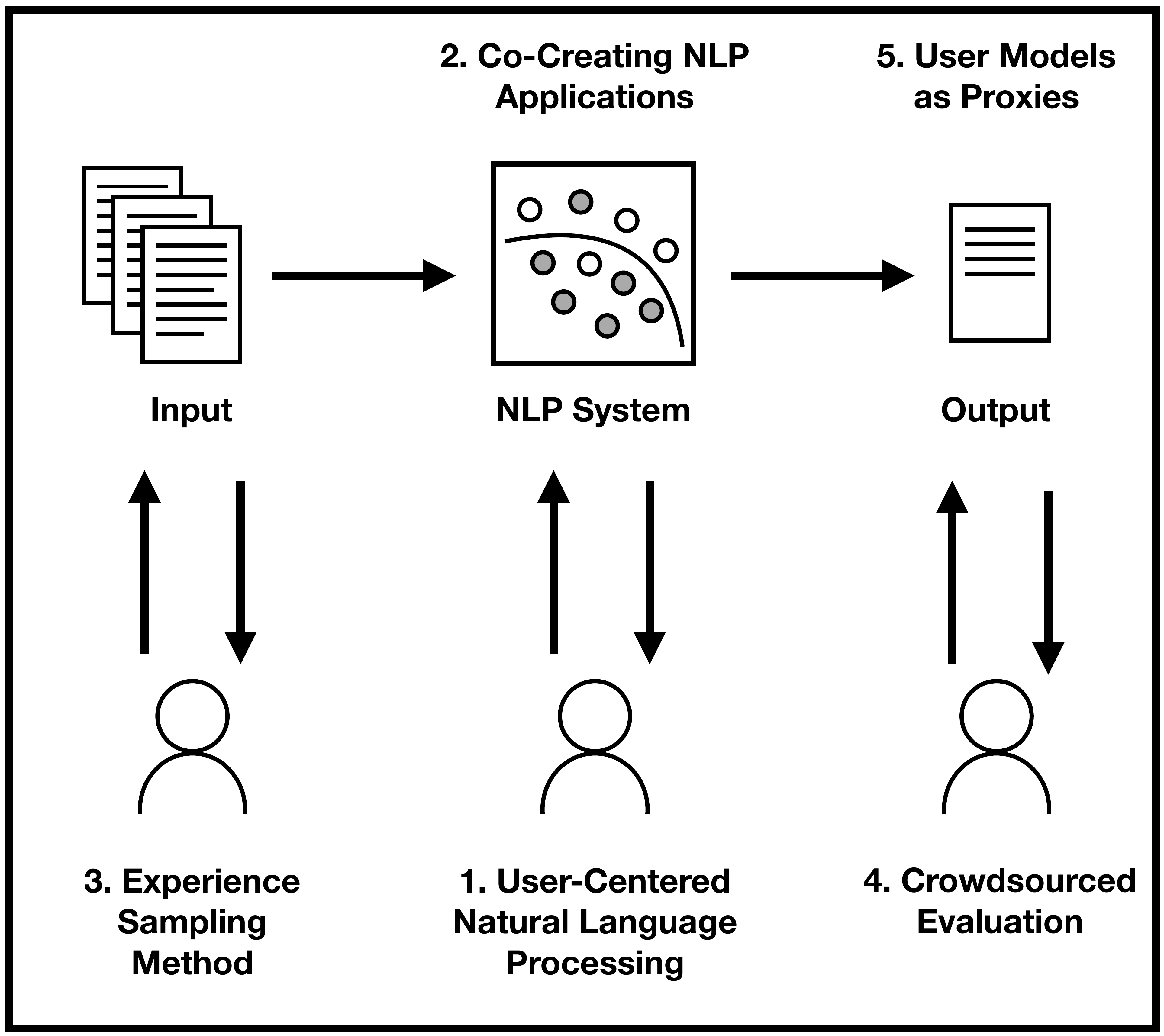}
\caption{The model situates the five methodological proposals in the context of an NLP system.}\label{fig:figure}
\end{figure}

Figure~\ref{fig:figure} situates the different methods in the context of HCI+NLP systems. The figure illustrates that two approaches are focused on the model side and three methods are focused on the user side. Methods 1 and 2 are focused on the NLP system itself. The \textit{1. User-Centered NLP} is at the heart of the model and focuses on users' understanding of the output and the explanations of the NLP system. While Method 2 is also strongly related to the user, we put it on the system side to highlight that when \textit{2. Co-Creating an NLP system}, the goal is not just to evaluate the experience with an NLP system, but to enable users to actively shape the system. This does not only include what the system looks like but means involving users in the problem formulation stage and allowing them to shape what problem is being solved. Considering the input that an NLP system is trained on, Method \textit{3. Experience Sampling} provides a simpler way of collecting metadata and more actively involving people in the collection of the dataset. Regarding the output of an NLP system, we showed the utility of \textit{4. Crowdsourcing} the Evaluation of NLP systems, which puts users into the loop to evaluate existing NLP systems at scale. The advantage of this is that a large number of users can be involved in the evaluation of the system. Finally, Method 5 proposes simulating real users through other ML-based systems. These \textit{5. User Models} can act as proxies for real users and allow a fast, automated evaluation of NLP systems at scale. We hope that this work informs novel approaches on how to standardize tools for large-scale interactive evaluations that will generate comparable and actionable benchmarks.

\section{Conclusion}

The five methods presented in Figure~\ref{fig:figure} cover the whole spectrum of HCI+NLP systems including the input, the NLP system, and the output of the system. Though each method has merits on its own, for successful future HCI+NLP applications, we believe that the whole will be greater than the sum of its parts. The design of future HCP+NLP applications should be centered around users (1) and involve them not only in the evaluation but also in the development and the problem formulation of an NLP system (2). Rich-meta data (3) that shapes the input of such a system are equally important as a thorough investigation of the output of the system, both by humans-in-the-loop (4) and by approaches based on computational methods that automate certain key aspects of such systems (5).

We hope that this overview of HCI and NLP methods is a useful starting point to engage interdisciplinary collaborations and to foster an exchange of what HCI and NLP have to offer each other methodologically. With this work, we hope to stimulate a discussion that brings HCI and NLP together and that advances the methodologies for technical and human-centered system design and evaluation in both fields.

\section{Acknowledgments}

This work was partially funded by the Deutsche Forschungsgemeinschaft (DFG, German Research Foundation) under project number 374666841, SFB 1342. This project is also partly funded by the Bavarian State Ministry of Science and the Arts and coordinated by the Bavarian Research Institute for Digital Transformation (bidt).

\bibliographystyle{acl_natbib}
\bibliography{bibliography}

\end{document}